\documentclass[10pt,twocolumn]{article}

\usepackage[margin=1in]{geometry}
\usepackage{graphicx}
\usepackage{booktabs}
\usepackage{amsmath}
\usepackage{hyperref}
\usepackage{xcolor}
\usepackage{multirow}
\usepackage{caption}
\usepackage{subcaption}
\usepackage{enumitem}
\usepackage[numbers]{natbib}
\usepackage{url}

\hypersetup{colorlinks=true, linkcolor=blue, citecolor=blue, urlcolor=blue}

\setlist{nosep, leftmargin=*}

\title{%
  A Synthetic Eye Movement Dataset for Script Reading Detection:\\
  Real Trajectory Replay on a 3D Eye Simulator%
}

\author{
  Kidus Zewde (kidus@scam.ai)\textsuperscript{1} \and
  Yuchen Zhou\textsuperscript{1} \and
  Dennis Ng\textsuperscript{1} \and
  Neo Tiangratanakul\textsuperscript{1} \and
  Tommy Duong\textsuperscript{1} \and
  Ankit Raj\textsuperscript{1} \and
  Yuxin Zhang\textsuperscript{1} \and
  Xingyu Shen\textsuperscript{1} \and
  Simiao Ren\textsuperscript{\dag} (benren@scam.ai)\thanks{Correspondence Author}
}

\date{}

\begin{document}
\maketitle
{\renewcommand{\thefootnote}{}\footnote{Authorship of this paper does not imply employment by or affiliation with Scam.AI (Reality Inc.); some authors are independent researchers who collaborated on this work.}\addtocounter{footnote}{-1}}

\vspace{-5ex}

\begin{abstract}
Large vision-language models have achieved remarkable capabilities by training on
massive internet-scale data, yet a fundamental asymmetry persists: while LLMs
can leverage self-supervised pretraining on abundant text and image data, the
same is not true for many behavioral modalities. Video-based behavioral data
--- gestures, eye movements, social signals --- remains scarce, expensive to
annotate, and privacy-sensitive. A promising alternative is \emph{simulation}:
replace real data collection with controlled synthetic generation to produce
automatically labeled data at scale.
We introduce infrastructure for this paradigm applied to eye movement,
a behavioral signal with applications across vision-language
modeling, virtual reality, robotics, accessibility systems, and cognitive
science. We present a pipeline for generating synthetic labeled eye movement
video by extracting real human iris trajectories from reference videos and
replaying them on a 3D eye movement simulator via headless browser automation.
Applying this to the task of script-reading detection during video interviews,
we release \texttt{final\_dataset\_v1}: 144 sessions (72\,reading, 72\,conversation)
totaling 12\,hours of synthetic eye movement video at 25\,fps.
Evaluation shows that generated trajectories preserve the
temporal dynamics of the source data (KS $D < 0.14$ across all metrics).
A matched frame-by-frame comparison
reveals that the 3D simulator exhibits bounded sensitivity at reading-scale
movements, attributable to the absence of coupled head movement --- a finding
that informs future simulator design.
The pipeline, dataset, and evaluation tools are released to support
downstream behavioral classifier development at the intersection of
behavioral modeling and vision-language systems.
\end{abstract}

\section{Introduction}
\label{sec:intro}

Large language and vision-language models have reached striking capabilities
by training on internet-scale data~\citep{brown2020gpt3, radford2021learning}.
A key enabling factor is the abundance of self-supervised training signal ---
text can be scraped at scale, and images can be paired with captions from alt-text
and surrounding context. Yet for many \emph{behavioral} modalities --- eye movements,
gestures, facial micro-expressions, social gaze patterns --- this data abundance
does not hold. Labeled behavioral video is scarce, expensive to annotate at scale,
and often privacy-sensitive. This asymmetry limits the development of models that
reason about human behavior from video.

A compelling approach to this data scarcity problem is \emph{behavioral simulation}:
rather than collecting infinite real recordings, one can simulate the behavioral
process in a controlled environment and generate labeled synthetic data at scale.
This paradigm has proven powerful in robotics (sim-to-real transfer) and computer
vision (synthetic training data for object detection), where generated scenes
can be randomized, labeled automatically, and scaled arbitrarily.
We take a step toward this question for eye movement: can a 3D eye simulator,
driven by real human iris trajectories, produce labeled behavioral video that
preserves the temporal dynamics of the source data?
We introduce the pipeline and dataset to enable this direction, and empirically
characterize what the simulation faithfully reproduces and where it falls short.

Eye movement is a particularly attractive behavioral signal for several reasons.
First, it is partially invariant to camera position and head orientation ---
the \emph{temporal pattern} (dynamics) of eye movement, rather than absolute gaze
direction, can distinguish behaviors without requiring per-subject geometric
calibration~\citep{hansen2010gaze}. Second, eye movements encode rich cognitive
state: reading patterns differ systematically from natural conversation; search
behaviors differ from comprehension; expertise is reflected in gaze
allocation~\citep{rayner1998eye, kunze2013know}. Third, gaze is a fundamental
signal across applications: training vision-language models on behavioral video,
gaze-contingent rendering in virtual and augmented reality, social robotics,
eye-based communication interfaces for accessibility, and cognitive science
research tools.

Applying this paradigm to a concrete setting, we focus on \emph{script reading
detection during video interviews}. Detecting whether a person is reading from
a script is practically useful for hiring, proctoring, and telehealth assessments.
Reading produces distinctive gaze dynamics --- structured horizontal sweeps with
regular fixations and saccades tracking text lines --- that differ from natural
conversational gaze, which tends to be center-biased with larger, less regular
movements. A system that can reliably detect reading from a webcam has broad
utility, yet the absence of labeled training data has made such systems
difficult to build and evaluate.

\textbf{Contributions.}
We introduce a pipeline and open dataset as a foundation for behavioral
eye movement simulation.
Our contributions are:
\begin{itemize}
  \item \textbf{A synthetic eye movement dataset} (\texttt{final\_dataset\_v1}):
    144 sessions, 12\,hours of labeled video generated by replaying real iris
    trajectories on a 3D eye simulator, with preserved temporal dynamics
    (KS $D < 0.14$ across speed, fixation, and saccade metrics).

  \item \textbf{A trajectory-to-simulator pipeline}: extracts iris landmarks
    from webcam video, calibrates to cursor-space, corrects resampling artifacts,
    and normalizes across subjects --- enabling faithful replay of real gaze
    patterns on a photorealistic eye model.

  \item \textbf{An empirical characterization of 3D eye simulator sensitivity}
    at reading-scale movements: a bounded amplitude limitation
    (30--42\% of real iris movement) explained by the absence of coupled
    head rotation, informing future simulator design.

  \item \textbf{An open release} of \texttt{final\_dataset\_v1}, the full
    generation pipeline, and evaluation tools to support downstream
    behavioral classifier development.
\end{itemize}
We evaluate distributional fidelity and report a key finding:
trajectory dynamics are well-preserved in the generated data, while the
simulator exhibits bounded visual sensitivity at reading scale --- a
characterization that informs both the use of this dataset and future
simulator improvements.
Whether the released data can successfully train a downstream behavioral
classifier is an open question left for future work.

\section{Related Work}
\label{sec:related}

\textbf{Simulation for training data in machine learning.}
The use of synthetic data to train perception models has a long history in
computer vision and robotics. Domain randomization, in which a simulator's
parameters are randomized over a large range, forces models to learn
domain-invariant features and enables sim-to-real transfer without requiring
photorealistic rendering~\citep{tobin2017domain}. Subsequent work demonstrated
that photorealistic synthetic data could match or exceed real data for tasks
including object detection, semantic segmentation, and depth
estimation~\citep{richter2016playing, dosovitskiy2017carla}.
More recently, the availability of large-scale rendered datasets has enabled
training of models for human pose estimation, face recognition, and hand
tracking with synthetic human models~\citep{varol2017learning, wood2021sim}.
These results establish the core premise of our work: synthetic data generated
by replaying real behavioral signals in a controlled simulator can serve as
a viable training data source.

\textbf{Deepfake and AI-generated content detection.}
The rise of generative AI has made synthetic media detection a critical
application area for simulation-based training data. Studies have shown that
multi-modal large language models can achieve competitive deepfake detection
performance with zero-shot generalization, sometimes surpassing traditional
pipelines on out-of-distribution data~\citep{ren2025multimodal}.
However, evaluations on real-world deployment reveal that many detectors
fail significantly under practical conditions, highlighting the importance
of benchmarking under realistic rather than curated test
sets~\citep{ren2025deepfake}.
A comprehensive zero-shot evaluation of 16 state-of-the-art AI-generated
image detection methods across 12 datasets found a 37 percentage-point gap
between the best and worst detectors, with training data alignment being the
critical factor for generalization~\citep{ren2026benchmark}.
Similar challenges arise in document fraud detection, where multi-modal LLMs
show promise but task-specific fine-tuning remains critical for reliable
performance~\citep{liang2025document}.
These results motivate the simulation paradigm we adopt: controlled
generation with known ground truth can bridge the gap between curated
benchmarks and real-world deployment.

\textbf{Gaze and eye movement as behavioral signal.}
Beyond gaze \emph{direction}, the temporal pattern of eye movement --- fixation
durations, saccade amplitudes, scanpath regularity --- encodes rich information
about cognitive state and task engagement. In controlled eye-tracking studies,
reading patterns reliably distinguish expert from novice readers~\citep{rayner1998eye},
and saccade planning has been modeled as an optimal trade-off between
information gathering and movement cost~\citep{salvucci2001cognitive}.
Reading detection from eye movement has been demonstrated in laboratory settings
using mobile eye trackers~\citep{kunze2013know}, and electrooculography-based
activity recognition has shown that eye movement patterns can classify tasks
without content-specific training~\citep{bulling2011recognition}.
However, these approaches require dedicated eye-tracking hardware that is
impractical for in-the-wild webcam settings.

\textbf{Appearance-based gaze estimation from video.}
Webcam-based gaze estimation has advanced rapidly with deep learning, enabling
 gaze estimation without specialized hardware. We surveyed 20 open-source
repositories and evaluated 7 that provided working pretrained
models (Table~\ref{tab:gaze_survey}). Among these, L2CS-Net performed best on accuracy and robustness in our setting
and is used in our fidelity evaluation (Section~\ref{sec:visual_fidelity}).

Synthetic eye images have also been used to augment training data for gaze
estimation itself~\citep{wood2015rendering, swirski2014rendering}, typically
via rendered 3D eyeball models with randomized illumination and identity
parameters. These approaches differ from ours: they use synthetic rendering to
produce training labels for a gaze \emph{estimator}, whereas we use a simulator
to produce behavioral \emph{video} driven by real trajectories for training a
behavioral \emph{classifier}.

\textbf{Behavioral data scarcity in vision-language models.}
Large vision-language models achieve strong performance partly because they can be
pretrained on internet-scale image-text pairs. For many behavioral modalities,
the same pretraining advantage does not apply: labeled behavioral video is
expensive to collect, annotation requires expertise, and personal data is
sensitive. The data scarcity is particularly acute for eye movement, where
existing datasets (e.g., GazeCapture~\citep{krafka2016eye}) require specialized
hardware or calibration, limiting scale. Generating synthetic behavioral data
by replaying real trajectories sidesteps these constraints, enabling dataset
generation at arbitrary scale with automatic labels.

\textbf{Cognitive models of eye movement during reading.}
Computational models of reading provide theoretical grounding for why gaze
dynamics differ between tasks. The E-Z Reader model~\citep{reichle2003ez}
proposes that fixation duration is determined by a familiarity-based lexical
processing cascade; the SWIFT model~\citep{engbert2005swift} treats saccade
planning as an emergent property of spatially distributed attention. These
models predict systematic differences in fixation and saccade statistics between
reading and non-reading gaze, which form the theoretical basis for dynamics-based
script detection. Our work complements these cognitive models by providing a
generative system that can produce realistic synthetic eye movement video at
scale, enabling machine learning approaches that build on these theoretical
predictions.

\textbf{Temporal smoothing for gaze signals.}
Raw gaze estimates from video contain significant frame-to-frame jitter due to
landmark localization noise. We employ EdgeGauss smoothing (our implementation) ---
a combination of heavy Gaussian filtering with edge-preserving saccade
restoration --- to produce clean temporal signals while preserving saccade
structure. This is a standard preprocessing step in gaze-based behavior analysis
and ensures that the extracted trajectories reflect genuine eye movement rather
than estimation noise.

\begin{table*}[t]
  \centering
  \caption{Survey of open-source gaze estimation methods. Seven
    models produced working inference with pretrained weights; 13
    were excluded for missing weights, wrong task, or incompatible
    frameworks.}
  \label{tab:gaze_survey}
  \small
  \begin{tabular}{@{}rlllll@{}}
    \toprule
    \textbf{\#} & \textbf{Model} & \textbf{Backbone}
      & \textbf{Accuracy} & \textbf{Params}
      & \textbf{GPU VRAM} \\
    \midrule
    1 & L2CS-Net   & ResNet-50 + RetinaFace
      & 3.92$^\circ$ MPII     & $\sim$24M + 29M det
      & 234\,MB \\
    2 & FAZE       & DenseNet DTED + MAML
      & $\sim$3.9$^\circ$ MPII & $\sim$6.9M
      & 30\,MB \\
    3 & GazeTR     & ResNet-18 + Transformer
      & SoTA ETH-XGaze        & $\sim$11.4M
      & 49\,MB \\
    4 & PureGaze   & ResNet-50 + Deconv
      & SoTA domain gen.      & $\sim$31.6M
      & 134\,MB \\
    5 & MobileGaze & ResNet-50 + RetinaFace
      & 11.33$^\circ$ Gaze360 & $\sim$24M + 29M det
      & 228\,MB \\
    6 & ptgaze     & ResNet-18
      & N/A (3 modes)         & $\sim$11.7M
      & 50\,MB \\
    7 & RT-GENE    & VGG16 + S3FD
      & SoTA MPII/RT-GENE     & $\sim$82M + S3FD
      & 557\,MB \\
    \midrule
    \multicolumn{6}{@{}l}{\textit{Excluded (13): wrong task (4),
      no weights (4), incompatible framework (3),
      incomplete (2)}} \\
    \bottomrule
  \end{tabular}
\end{table*}

\section{Methodology}
\label{sec:method}

\begin{figure*}[t]
  \centering
  \includegraphics[width=\textwidth]{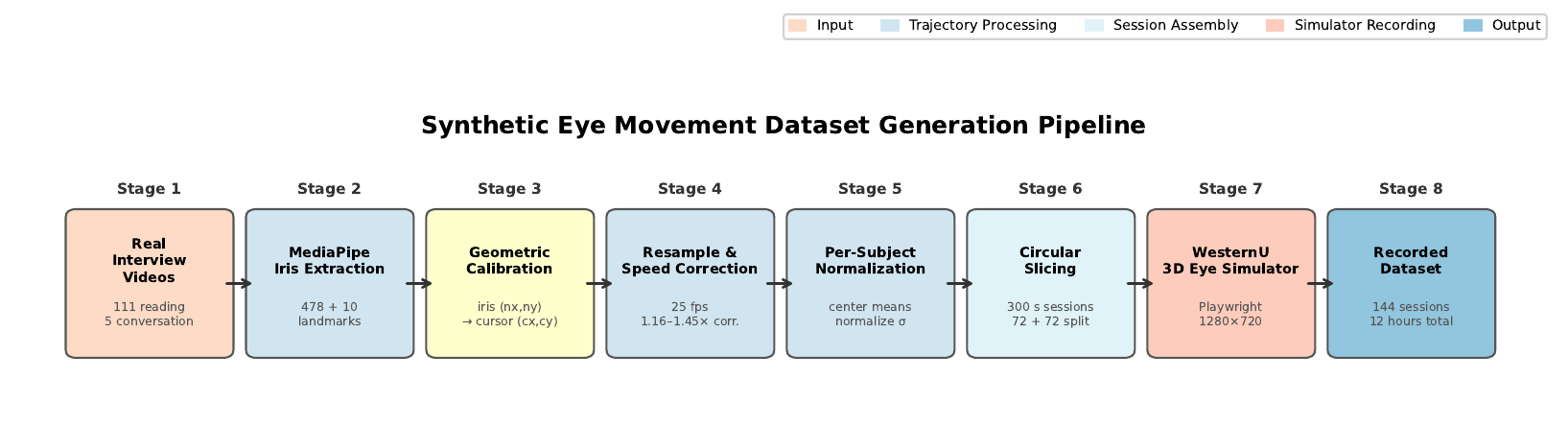}
  \caption{Generation pipeline overview. Real interview videos are processed
    through six trajectory-processing stages and finally recorded on the
    WesternU 3D Eye Simulator via Playwright automation.}
  \label{fig:pipeline}
\end{figure*}

Figure~\ref{fig:pipeline} shows the full pipeline.
We describe each stage below.

\subsection{Stage 1: Source Trajectory Extraction}
\label{sec:iris_extraction}

\begin{figure}[t]
  \centering
  \includegraphics[width=0.75\linewidth]{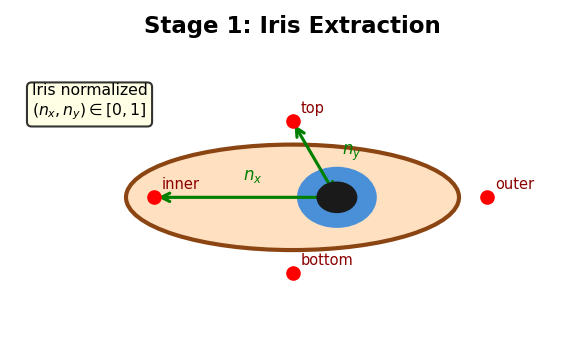}
  \caption{Stage 1: Iris landmarks (red dots) define the eye aperture;
    normalized iris position $(n_x, n_y)$ is computed relative to corner
    and eyelid landmarks.}
  \label{fig:stage1}
\end{figure}

We extract per-frame iris position from reference interview videos using
MediaPipe FaceLandmarker~\cite{mediapipe2019}, which detects 478 facial
landmarks plus 10 iris-specific landmarks per face.
As illustrated in Figure~\ref{fig:stage1}, the iris center is normalized
relative to the eye corner and eyelid landmarks to produce coordinates
$(n_x, n_y) \in [0, 1]$; left and right eye values are averaged to reduce
per-eye noise.  Full normalization details are given in Appendix~\ref{app:iris}.

\textbf{Source data.}
We processed 11 reading interview videos (subjects reading from a screen during
a video call) and 5 natural conversation videos, yielding 16 trajectory files
totaling approximately 310{,}000 source frames.
The reading videos span 11 distinct subjects; the conversation videos span
a range of identities and discussion topics.
All reference videos were collected internally from consenting participants
(see Ethics Statement).

\subsection{Stage 2: Geometric Calibration}
\label{sec:calibration}

\begin{figure}[t]
  \centering
  \includegraphics[width=0.75\linewidth]{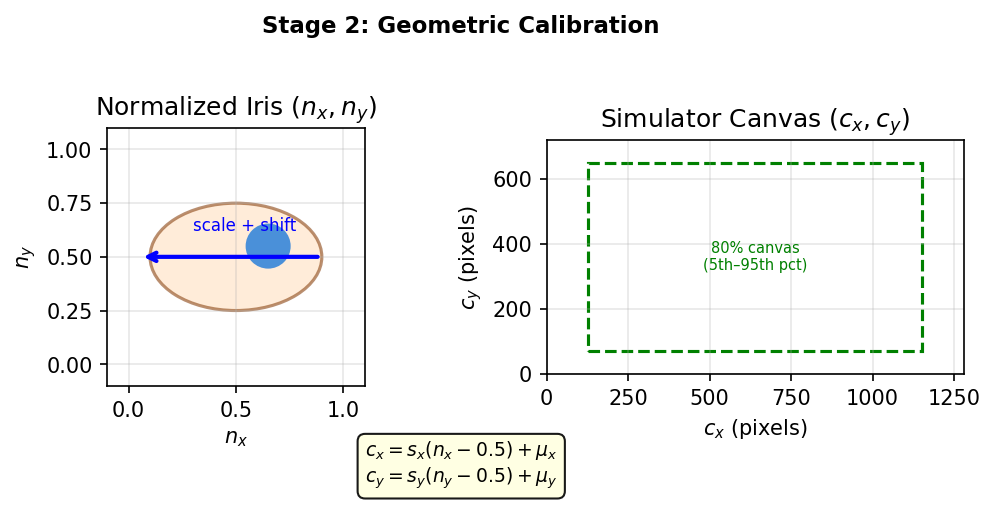}
  \caption{Stage 2: Normalized iris coordinates $(n_x, n_y) \in [0,1]$ are
    mapped to cursor position $(c_x, c_y)$ on the 1280$\times$720\,px simulator
    canvas via scale and shift.}
  \label{fig:stage2}
\end{figure}

We map normalized iris coordinates to simulator canvas coordinates
(1280$\times$720\,px) via a linear scale-and-shift model
(Figure~\ref{fig:stage2}; Appendix~\ref{app:calibration}):
$(c_x, c_y) = (s_x, s_y) \odot (n_x - 0.5, n_y - 0.5) + (\mu_x, \mu_y)$.
In auto-scale mode, $(s_x, s_y)$ and $(\mu_x, \mu_y)$ are derived from the
5th--95th percentile range of observed iris values, mapping to 80\% of the
canvas extent.
Blink artifacts are filtered via a 3$\times$IQR outlier rule and forward-filled.

\subsection{Stage 3: Resampling and Speed Correction}
\label{sec:resample}

\begin{figure}[t]
  \centering
  \includegraphics[width=0.8\linewidth]{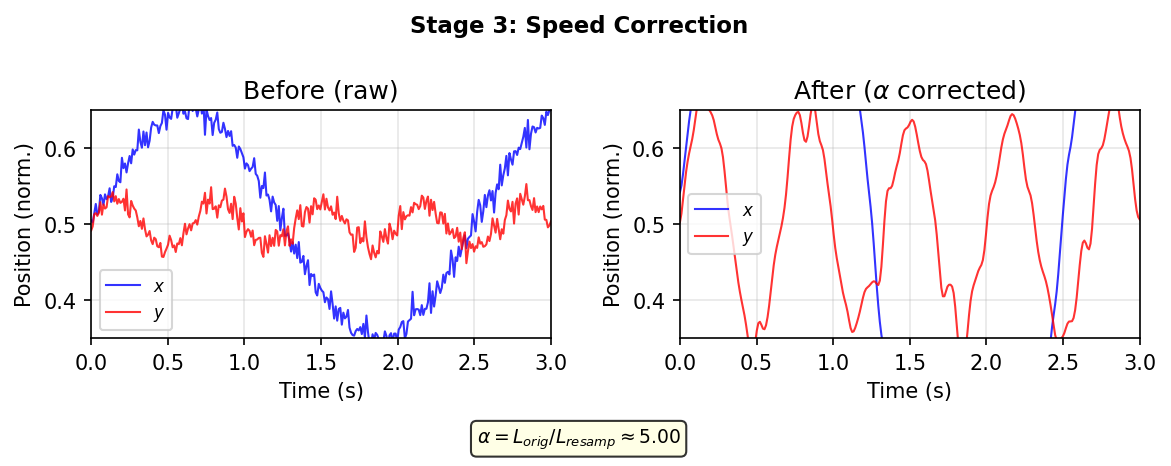}
  \caption{Stage 3: Speed correction restores trajectory path length after
    resampling.  $\alpha \approx 1.16$--$1.45\times$ amplifies the smoothed
    signal back to the original travel distance.}
  \label{fig:stage3}
\end{figure}

Source videos have variable native frame rates (14.96--30.03\,fps).
We resample all trajectories to 25\,fps via per-frame linear interpolation.
Because interpolation smooths high-frequency jitter and reduces total path
length, we apply a post-resampling speed correction
(Figure~\ref{fig:stage3}; Appendix~\ref{app:speed}):
$\mathbf{x}' = \bar{\mathbf{x}} + \alpha(\mathbf{x} - \bar{\mathbf{x}})$,
where $\alpha = L_{\text{orig}}/L_{\text{resamp}} \approx 1.16$--$1.45\times$.

\subsection{Stage 4: Per-Subject Normalization and Concatenation}
\label{sec:normalization}

\begin{figure}[t]
  \centering
  \includegraphics[width=0.75\linewidth]{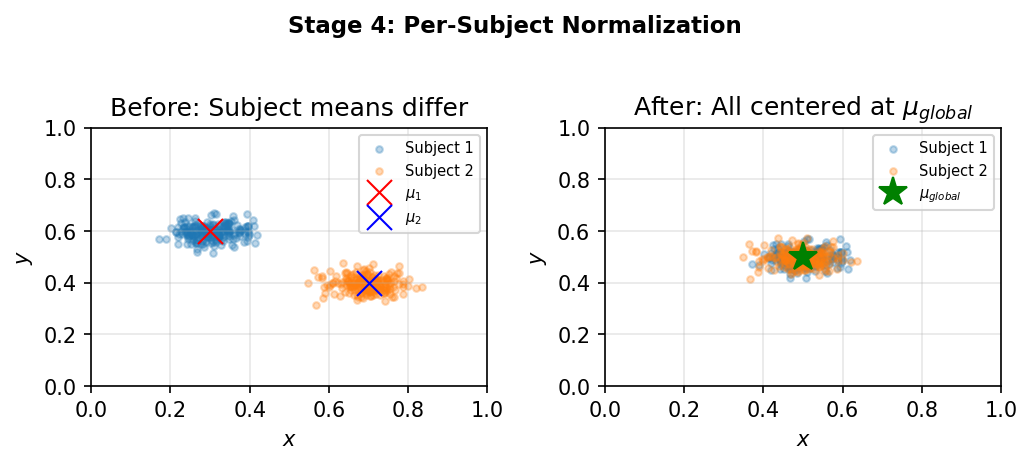}
  \caption{Stage 4: Per-subject normalization centers each subject's
    distribution on the global mean, removing inter-subject baseline offsets.}
  \label{fig:stage4}
\end{figure}

Different subjects have different resting gaze positions.
Before concatenation, we shift each video's mean to a common global center
(Figure~\ref{fig:stage4}; Appendix~\ref{app:normalize}):
$\mathbf{x}^{(v)} \leftarrow \mathbf{x}^{(v)} +
(\boldsymbol{\mu}_{\text{global}} - \boldsymbol{\mu}^{(v)})$.
All normalized trajectories of the same class are then concatenated into a
single long array.

\subsection{Stage 5: Circular Slicing and Speed Scaling}
\label{sec:slicing}

\begin{figure}[t]
  \centering
  \includegraphics[width=0.8\linewidth]{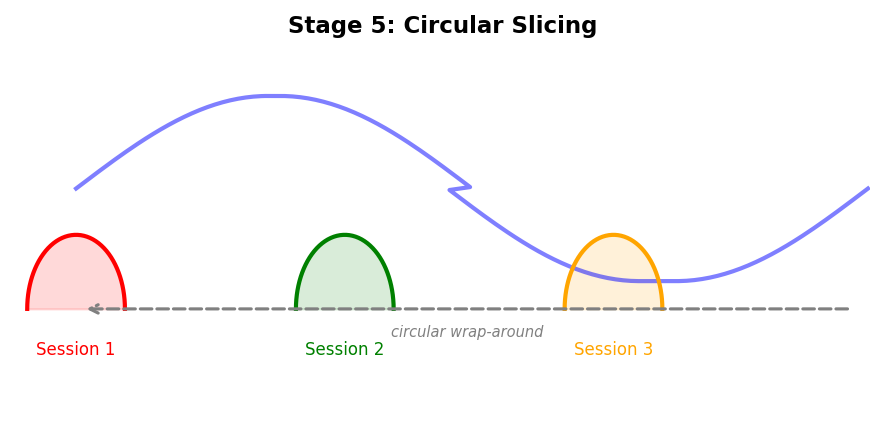}
  \caption{Stage 5: Circular slicing extracts sessions at uniform stride
    from the concatenated trajectory.  Wrap-around (dashed arrow) enables
    uniform coverage from a finite array.}
  \label{fig:stage5}
\end{figure}

From the concatenated array, we extract $N$ sessions of $F$ frames each
(default: $N{=}72$, $F{=}7500$, i.e., 300\,s at 25\,fps).
Sessions are distributed uniformly with stride $\lfloor T/N \rfloor$ using
modular (circular) indexing
(Figure~\ref{fig:stage5}; Appendix~\ref{app:slicing}):
$\text{idx}_j = (i\cdot\lfloor T/N\rfloor + j) \bmod T$.
Because circular wrap-around reuses the same underlying trajectory frames,
sessions are not fully independent (see Section~\ref{sec:dist_fidelity}).
A per-class speed scale $\alpha_{\text{reading}} = 1.05$ and
$\alpha_{\text{conv}} = 1.0$ is applied before slicing; the reading
scale was set empirically to partially compensate for the simulator's
reduced amplitude response at reading-scale inputs
(Section~\ref{sec:sim_limitation}).

\subsection{Stage 6: 3D Eye Simulator and Recording}
\label{sec:simulator}

\begin{figure}[t]
  \centering
  \includegraphics[width=0.65\linewidth]{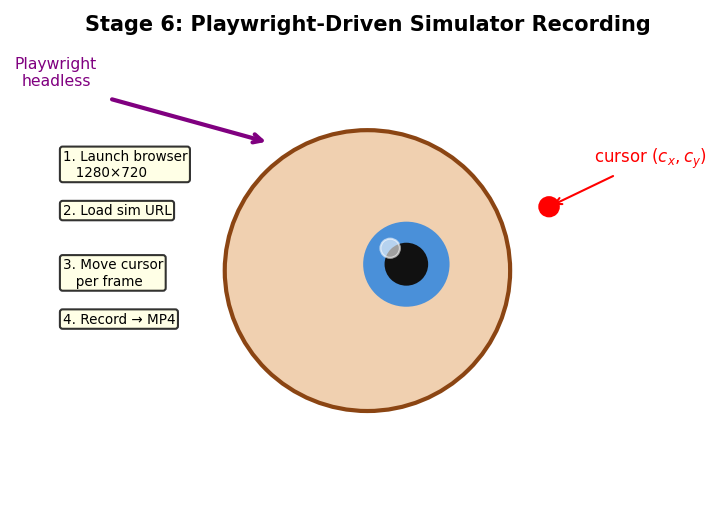}
  \caption{Stage 6: Playwright headless browser automation drives the WesternU
    3D eye simulator by moving the cursor to each trajectory position and
    recording the resulting eye video.}
  \label{fig:stage6}
\end{figure}

We use the WesternU 3D Eye Movement Simulator~\cite{westernu_eye_sim},
a Unity WebGL application whose gaze direction follows the mouse cursor.
Recording is automated via Playwright~\cite{playwright2023} headless browser
control (Figure~\ref{fig:stage6}):

\begin{enumerate}
  \item Launch headless Chromium with a 1280$\times$720 viewport
    and video recording enabled.
  \item Navigate to the simulator URL and wait for Unity
    initialization.
  \item For each frame: move the cursor to the trajectory
    position, then wait $1000/\text{fps}$\,ms.
  \item On completion, close the browser context and convert the
    WebM recording to MP4 via FFmpeg.
\end{enumerate}

\section{Dataset Description}
\label{sec:dataset}

Table~\ref{tab:dataset} summarizes the dataset statistics.

\begin{table}[t]
  \centering
  \caption{Dataset statistics for \texttt{final\_dataset\_v1}.}
  \label{tab:dataset}
  \small
  \begin{tabular}{@{}lp{3.0cm}@{}}
    \toprule
    \textbf{Property} & \textbf{Value} \\
    \midrule
    Sessions        & 144 (72\,R + 72\,C) \\
    Duration/session & 300\,s (5\,min) \\
    Frame rate       & 25\,fps \\
    Resolution       & 1280$\times$720 \\
    Total duration   & 12\,hours \\
    Total size       & 1.7\,GB \\
    Reading sources  & 11 subjects \\
    Conv.\ sources   & 5 videos \\
    Speed scale (R/C) & 1.05\,/\,1.0 \\
    \bottomrule
  \end{tabular}
\end{table}

\textbf{Per-session output.}
Each session consists of an MP4 video and a paired trajectory CSV with columns
\texttt{frame}, \texttt{x}, \texttt{y} (cursor position in canvas pixels).
A \texttt{labels.csv} provides binary labels (1 = reading, 0 = conversation)
and \texttt{metadata.json} records all generation parameters for
reproducibility.

\textbf{Reading vs.\ conversation characteristics.}
Reading trajectories exhibit structured horizontal sweeps (tracking text lines),
regular fixation--saccade patterns, and relatively small spatial extent
($\sim$8\% of canvas width).
Conversation trajectories are center-biased with larger, irregular movements
spanning a wider spatial range ($\sim$20\% of canvas width).

\section{Evaluation}
\label{sec:evaluation}

We evaluate two aspects of dataset quality: (1)~whether the generated
trajectories preserve the distributional properties of the source reference
data, and (2)~how faithfully the simulator maps cursor input to visible iris
movement.

\textbf{Note on session independence.} Because sessions are extracted from a
concatenated trajectory via circular slicing (Equation~\ref{app:eq:circular_slicing}),
sessions share overlapping frames from the same underlying source trajectory.
Consequently, the generated sessions are not statistically independent, and the
distributional fidelity comparisons below reflect how well the session pool
captures source dynamics, not independent draws from a novel distribution.

\subsection{Distributional Fidelity}
\label{sec:dist_fidelity}

We compare the statistical properties of generated trajectory CSVs against
their source reference trajectories across three key metrics:
\begin{itemize}
  \item \textbf{Speed}: frame-to-frame Euclidean displacement (px/frame).
  \item \textbf{Fixation duration}: using an I-VT (velocity-threshold) detector
    with threshold 5\,px/frame; fixations are contiguous runs of frames below
    threshold, and duration is measured in frames.
  \item \textbf{Saccade amplitude}: I-VT with threshold 15\,px/frame; saccades
    are contiguous runs above threshold, and amplitude is total Euclidean
    displacement during the run (px).
\end{itemize}

\begin{figure*}[t]
  \centering
  \includegraphics[width=\textwidth]{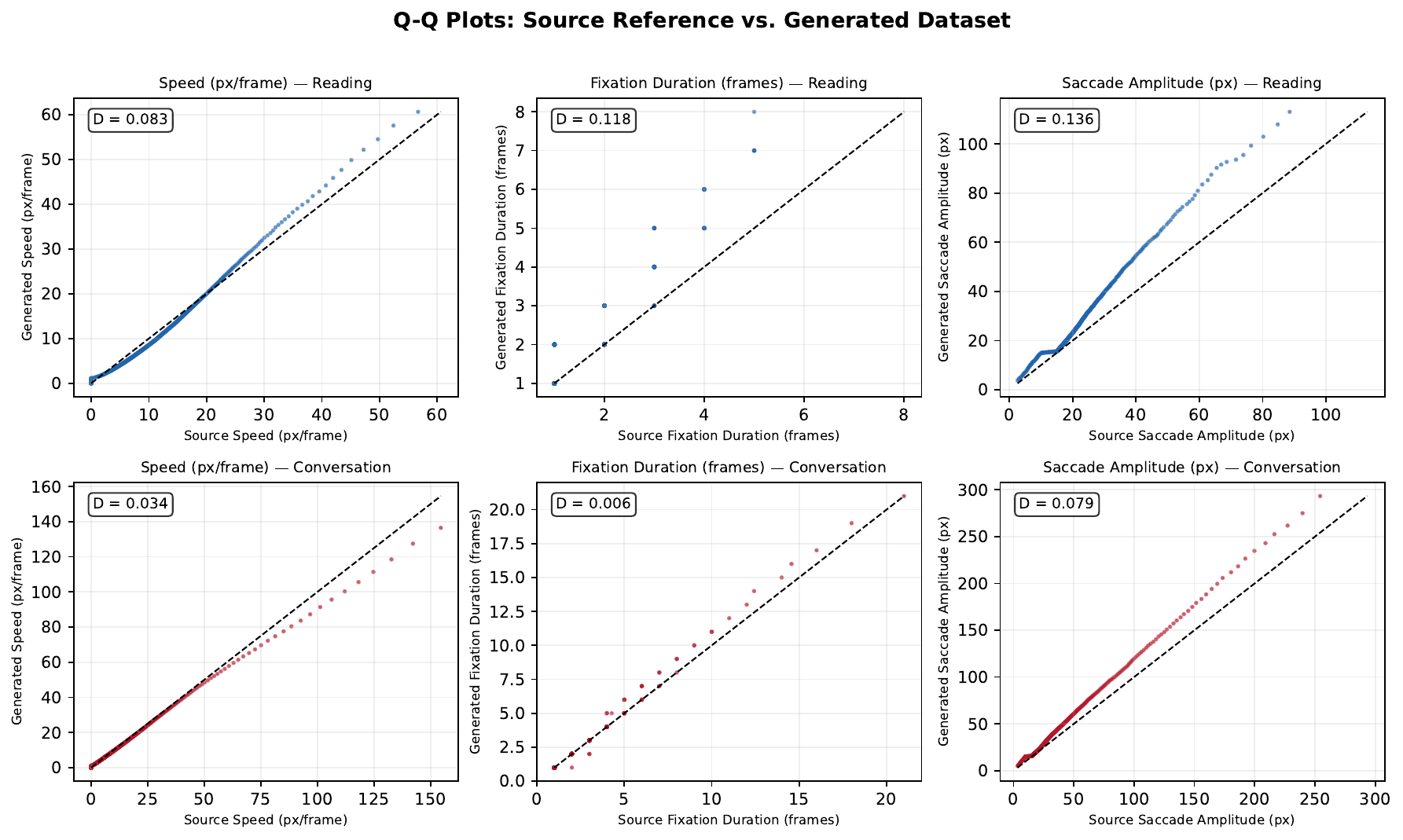}
  \caption{Q-Q plots: source vs.\ generated quantiles for speed, fixation
    duration, and saccade amplitude, across reading and conversation classes.
    Each panel shows generated quantiles (Y-axis) against source quantiles
    (X-axis); the dashed line is the 1:1 reference.
    KS $D$-statistics are annotated in each panel.}
  \label{fig:distributions}
\end{figure*}

Figure~\ref{fig:distributions} shows Q-Q plots comparing source and generated
distributions for each metric and class.  Points lying on the 1:1 diagonal
indicate perfect distributional match; deviations indicate systematic shift.
The generated data closely matches the source for both reading and conversation
classes, with KS $D$-statistics ranging from 0.006 (conversation fixation
duration) to 0.136 (reading saccade amplitude).

Two-sample Kolmogorov--Smirnov tests quantify the distributional match
(Table~\ref{tab:ks}).
For conversation, $D = 0.006$--$0.079$ indicates excellent distributional
match. For reading, $D = 0.083$--$0.136$ is notably larger; this is
consistent with the higher reuse ratio for reading (72 sessions from
33\,min of source data) compared to conversation (72 sessions from
183\,min), which introduces more repeated circular-wrapped patterns.
The $p$-values are small throughout due to the large sample sizes
($>$45K source frames), which is expected; they do not indicate practical
significance on their own. The $D$ statistics themselves indicate
acceptable distributional fidelity for both classes, with reading showing
greater deviation than conversation, likely reflecting the higher reuse
ratio rather than pipeline error.

\begin{table}[t]
  \centering
  \caption{Two-sample KS test statistics comparing source vs.\ generated
    distributions. Lower $D$ indicates better distributional match.}
  \label{tab:ks}
  \begin{tabular}{@{}lcc@{}}
    \toprule
    \textbf{Metric} & \textbf{$D$ (Reading)} & \textbf{$D$ (Conv.)} \\
    \midrule
    Speed (px/frame)       & 0.083 & 0.034 \\
    Fixation duration      & 0.118 & 0.006 \\
    Saccade amplitude      & 0.136 & 0.079 \\
    \bottomrule
  \end{tabular}
\end{table}

\begin{figure*}[t]
  \centering
  \includegraphics[width=\textwidth]{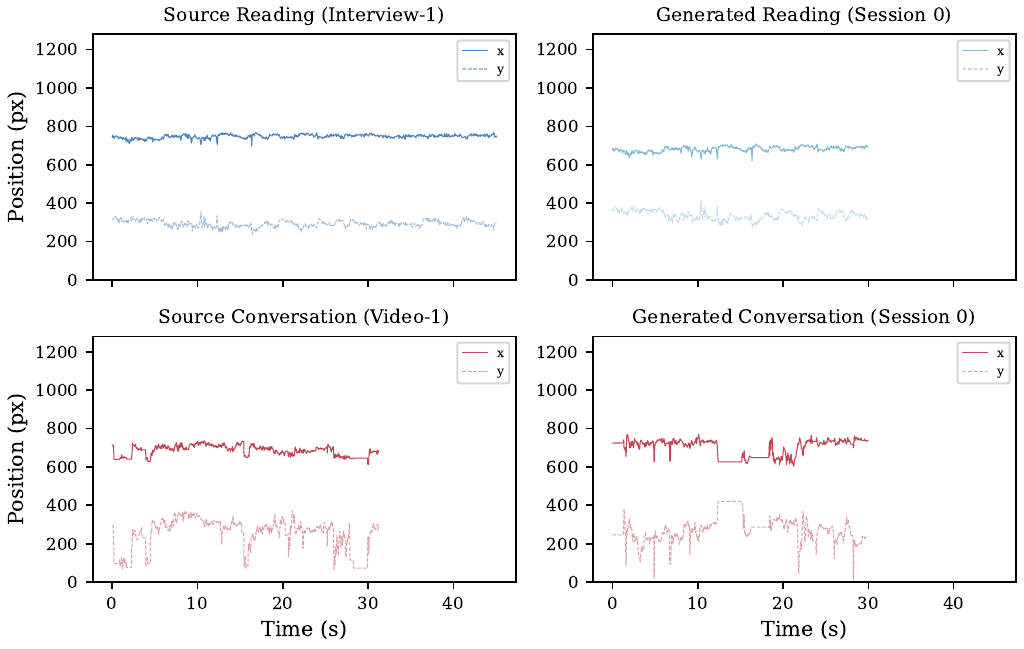}
  \caption{Trajectory time-series (30-second windows).
    Left: source reference trajectories from real interview videos.
    Right: corresponding generated sessions.
    The generated trajectories preserve the characteristic temporal structure:
    structured horizontal sweeps for reading, irregular center-biased movement
    for conversation.}
  \label{fig:trajectories}
\end{figure*}

Figure~\ref{fig:trajectories} shows representative time-series windows
comparing source and generated trajectories.

\subsection{Visual Mapping Fidelity}
\label{sec:visual_fidelity}

Beyond trajectory preservation, we evaluate how faithfully the simulator
translates cursor input into visible iris movement.
We conduct a \emph{matched apple-to-apple comparison} as a case study on one
reading subject (Tudimilla): we take this subject's exact MediaPipe-extracted
iris trajectory, convert it to cursor coordinates using the same calibration
pipeline, replay it on the simulator, then extract the simulator's resulting
iris positions and compare frame-by-frame against the original.
While this case study represents a single subject, the subject exhibits
representative reading behavior (structured horizontal sweeps) and was selected
as a representative case of reading-scale movement on the simulator.

\textbf{Quantitative results} (Table~\ref{tab:sim_fidelity}).
The mean normalized iris position error is 0.053 (in $[0, 1]$ space), and the
simulator's iris movement amplitude is 30--42\% of the real subject's amplitude.
Temporal correlation between cursor input and simulator iris output is near zero
($r_x = 0.006$, $r_y = 0.077$).

\begin{table}[t]
  \centering
  \caption{Matched iris comparison metrics (Tudimilla, 120\,s, 3000 frames).}
  \label{tab:sim_fidelity}
  \begin{tabular}{@{}ll@{}}
    \toprule
    \textbf{Metric} & \textbf{Value} \\
    \midrule
    Mean iris error (norm.)     & 0.053 \\
    Median iris error           & 0.036 \\
    95th percentile error       & 0.161 \\
    Correlation (X / Y)         & 0.006 / 0.077 \\
    Amplitude ratio (X / Y)     & 0.30 / 0.42 \\
    Cursor range (X)            & 100\,px (7.8\% of 1280) \\
    Cursor range (Y)            & 155\,px (21.5\% of 720) \\
    \bottomrule
  \end{tabular}
\end{table}

\begin{figure}[t]
  \centering
  \includegraphics[width=\columnwidth]{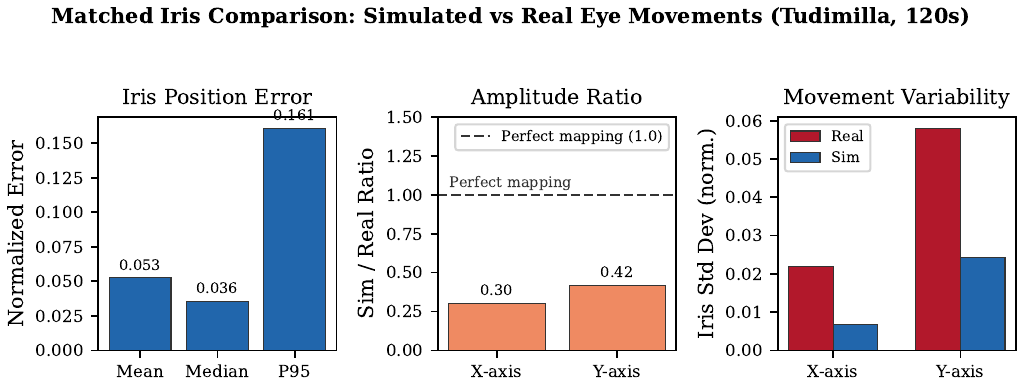}
  \caption{Sim fidelity metrics from matched comparison.
    Left: iris position error statistics.
    Center: amplitude ratio (sim/real).
    Right: movement variability (std dev) comparison.
    The simulator produces 30--42\% of the real subject's iris amplitude.}
  \label{fig:sim_fidelity}
\end{figure}

\begin{figure*}[t]
  \centering
  \includegraphics[width=\textwidth]{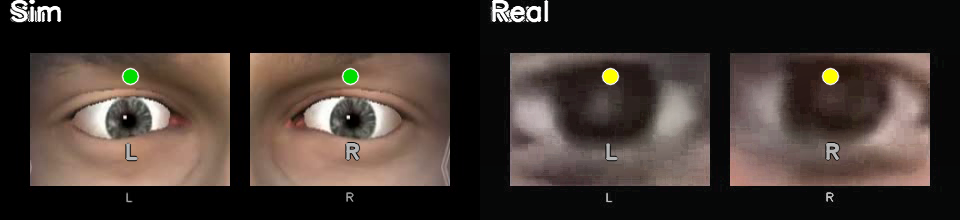}
  \caption{Apple-to-apple qualitative comparison at $t{=}15$\,s.
    Left: simulator replay (dark background, eye pinned).
    Right: original human reference video.
    Both eyes are normalized to a common center; only iris movement is visible.
    The simulator preserves the temporal pattern of the source trajectory
    with reduced visible amplitude at this scale.}
  \label{fig:apple_to_apple}
\end{figure*}

\subsection{Simulator Sensitivity: A Scale Limitation}
\label{sec:sim_limitation}

The low correlation and reduced amplitude in Table~\ref{tab:sim_fidelity}
reflect a fundamental characteristic of the simulation setup rather than a
pipeline error.
Three factors contribute:

\begin{enumerate}
  \item \textbf{Reading eye movements are inherently small.}
    During reading, eye-in-head rotation spans approximately 8\% of the
    simulator's 1280-pixel canvas width.
    The simulator, designed as an anatomy teaching tool, maps cursor position
    across the full canvas to eye rotation---at reading-scale inputs, the
    iris response is near the simulator's noise floor.

  \item \textbf{Absence of head movement.}
    In natural reading, head movement contributes 60--70\% of the total gaze
    sweep~\cite{freedman2008coordination}.
    The simulator receives only the eye-in-head rotation component (30--40\%
    of the full movement), further reducing the effective input range.

  \item \textbf{Conversation movements are well-captured.}
    Conversation gaze movements are larger and more varied, falling within the
    simulator's responsive range.
    This creates a natural separation between reading and conversation patterns
    in the generated data, which is preserved despite the reduced reading
    amplitude.
\end{enumerate}

\textbf{Implication for the dataset.}
The generated trajectories faithfully preserve the \emph{temporal dynamics}
(speed, fixation patterns, saccade structure) of the source data.
The visual mapping introduces bounded attenuation at reading scale, but the
critical class-distinguishing signal---the difference in temporal patterns
between reading and conversation---remains intact.

\section{Discussion}
\label{sec:discussion}

\textbf{Failed approaches.}
Before arriving at the current pipeline, we explored several alternatives:
\begin{itemize}
  \item \emph{Synthetic parametric patterns}: Programmatically generated
    fixation--saccade sequences produced unrealistic extreme eye angles and
    failed to capture real reading dynamics.
  \item \emph{GPU pixel matching}: Matching real video frames to a pre-rendered
    simulator grid via L2 pixel distance collapsed to a small set of duplicate
    positions---pixel-level appearance matching proved fundamentally unsuitable.
  \item \emph{Ornstein--Uhlenbeck process}: Synthetic conversation trajectories
    via OU drift were not calibrated from real data and were replaced by real
    trajectory replay.
\end{itemize}
The common lesson is that \emph{real trajectory replay} succeeds where
parametric generation fails, because real iris trajectories carry the full
complexity of human eye movement dynamics.

\textbf{Limitations.}
The simulator uses a single eye model without head movement, bounding visual
fidelity at reading scale (Section~\ref{sec:sim_limitation}).
The reading class draws from 11 subjects while the conversation class draws
from 5 videos (183\,min), creating an asymmetric reuse ratio that accounts
for the higher KS $D$ values observed for reading.
The dataset currently provides cursor trajectories and rendered eye video but
not head pose or full-face video.
Critically, this paper does not train or evaluate a downstream behavioral
classifier on \texttt{final\_dataset\_v1}; whether the generated data can
successfully train such a model remains an open question and the primary
direction for future work.

\textbf{Future work.}
The most important next step is training and evaluating a behavioral
classifier (e.g., a 1D-CNN on per-session trajectory features) on
\texttt{final\_dataset\_v1} and testing on real interview video, to directly
validate whether the pipeline generates useful training data.
A head-movement-aware simulator (e.g., a full face model with coupled
eye--head rotation) would significantly improve visual fidelity for
reading-scale movements.
Trajectory amplification---scaling reading cursor movements to fill more of the
simulator's responsive range---is a simpler near-term mitigation.
Multi-identity eye models would add visual diversity across sessions.

\section{Conclusion}
\label{sec:conclusion}

We presented a pipeline and open dataset for generating synthetic labeled eye
movement video by replaying real human iris trajectories on a 3D eye simulator.
The resulting dataset, \texttt{final\_dataset\_v1}, provides 12~hours of
balanced reading and conversation eye movement data at 25\,fps.
Distributional evaluation shows that the generated trajectories preserve
the speed, fixation, and saccade characteristics of the source data
(KS $D < 0.14$).
A matched frame-by-frame comparison reveals a simulator sensitivity limitation
at reading scale, attributable to the absence of head movement---a finding
that informs future simulator design.
The dataset, generation pipeline, and evaluation tools are released openly
to enable downstream behavioral classifier development and further
exploration of eye movement simulation for vision-language applications.


\appendix
\section{Mathematical Formalism}
\label{app:formalism}

\subsection{Iris Normalization}
\label{app:iris}

For each video frame, MediaPipe FaceLandmarker provides 478 facial landmarks
plus 10 iris landmarks.  The normalized iris position is computed as:
\begin{equation}
  n_x = \frac{x_{\text{iris}} - x_{\text{inner}}}{x_{\text{outer}} - x_{\text{inner}}}, \qquad
  n_y = \frac{y_{\text{iris}} - y_{\text{top}}}{y_{\text{bottom}} - y_{\text{top}}}
  \label{app:eq:iris_norm}
\end{equation}
where $(n_x, n_y) \in [0, 1]$ are the normalized horizontal and vertical
coordinates within the eye aperture, respectively.  Left and right eye
values are averaged per frame to reduce per-eye noise.

\subsection{Geometric Calibration}
\label{app:calibration}

Normalized iris coordinates are mapped to simulator canvas coordinates
$(c_x, c_y)$ (1280$\times$720\,px) via a linear geometric model:
\begin{equation}
  c_x = s_x \cdot (n_x - 0.5) + \mu_x, \qquad
  c_y = s_y \cdot (n_y - 0.5) + \mu_y
  \label{app:eq:calibration}
\end{equation}
where $(s_x, s_y)$ are per-axis scale factors and $(\mu_x, \mu_y)$ are
center offsets.  In auto-scale mode, $(s_x, s_y)$ map the 5th--95th
percentile range of observed $(n_x, n_y)$ to 80\% of the canvas extent.

\textbf{Blink filtering.}  When the eyelids close, the iris normalization
(Equation~\ref{app:eq:iris_norm}) produces outlier values (spikes or
collapses).  Blink artifacts are detected via a 3$\times$IQR rule on each
axis independently and replaced by forward-fill from the last valid value.

\subsection{Resampling and Speed Correction}
\label{app:speed}

Source videos at variable frame rates $f_v \in [14.96, 30.03]$\,fps are
resampled to the target rate $f_{\text{tgt}} = 25$\,fps via per-frame linear
interpolation of $(c_x, c_y)$.

Resampling acts as a low-pass filter and shortens the total path length.
To restore the original travel distance, we apply a post-resampling speed
correction:
\begin{equation}
  \mathbf{x}'_i = \bar{\mathbf{x}} + \alpha\,(\mathbf{x}_i - \bar{\mathbf{x}}), \qquad
  \alpha = \frac{L_{\text{orig}}}{L_{\text{resamp}}}
  \label{app:eq:speed_correction}
\end{equation}
where $\mathbf{x}_i = (c_x, c_y)$ at frame $i$, $\bar{\mathbf{x}}$ is the
trajectory centroid, and $L = \sum_i \|\mathbf{x}_{i+1} - \mathbf{x}_i\|_2$
is the Euclidean path length.  Typical values are
$\alpha \in [1.16, 1.45]$.

\subsection{Per-Subject Normalization}
\label{app:normalize}

Different subjects have different resting gaze baselines.  Before
concatenation, each video $v$ is centered on the global mean
$\boldsymbol{\mu}_{\text{global}}$:
\begin{equation}
  \mathbf{x}_i^{(v)} \leftarrow \mathbf{x}_i^{(v)}
    + \bigl(\boldsymbol{\mu}_{\text{global}} - \boldsymbol{\mu}^{(v)}\bigr)
  \label{app:eq:normalization}
\end{equation}
where $\boldsymbol{\mu}^{(v)}$ is the mean cursor position of video $v$.
After normalization, all trajectories of the same class are concatenated into
a single long array.

\subsection{Circular Slicing}
\label{app:slicing}

From a concatenated array of length $T$, we extract $N$ sessions of $F$
frames each (default $N{=}72$, $F{=}7500$ for 300\,s at 25\,fps) by
distributing sessions uniformly with stride $\lfloor T/N\rfloor$ and using
modular indexing for circular wrap-around:
\begin{equation}
  \text{idx}_j = \bigl(i \cdot \lfloor T/N\rfloor + j\bigr) \bmod T,
  \quad j = 0, \ldots, F{-}1
  \label{app:eq:circular_slicing}
\end{equation}
for session $i = 0, \ldots, N{-}1$.  The modular operation reuses earlier
frames when the stride reaches the end of the array.

A per-class speed scale is applied before slicing:
$\mathbf{x} \leftarrow \bar{\mathbf{x}} + \alpha(\mathbf{x} - \bar{\mathbf{x}})$,
with $\alpha_{\text{reading}} = 1.05$ and $\alpha_{\text{conv}} = 1.0$.

\section*{Ethics Statement}

All reference videos used for iris trajectory extraction were collected
internally from employees who provided informed consent for their recordings
to be used in research under an internal data collection consent process.
No personally identifiable information (PII) beyond iris position coordinates
is retained in the released dataset; the source videos themselves are not
released.
The synthetic eye movement videos in \texttt{final\_dataset\_v1} are
generated from a 3D eye simulator and contain no real participant footage.

\section*{Data and Code Availability}
Supporting code and the non-restricted resources for this work are available from the Scam.AI
Research portal, \url{https://scam.ai/research}~\cite{scamai_research}, with document- and
PDF-forgery resources maintained at Trustable, \url{https://trustable.ai}~\cite{trustable_ai}.
As noted in the paper, the raw source videos used in this study are not publicly released; requests
and further details are handled through the Scam.AI Research portal.

\end{document}